\documentclass[sigconf, screen, dvipsnames, nonacm]{acmart}

\usepackage{hyperref}
\usepackage[noend]{algorithmic}
\usepackage{algorithm}
\usepackage{multirow}
\usepackage{amsmath}
\usepackage{bbm}
\usepackage{booktabs}
\usepackage[inline]{enumitem} % Inline lists
\usepackage{subcaption}
\usepackage{bm} % Bold greek
\usepackage{xcolor}
\usepackage{balance}
\usepackage{lipsum}

\usepackage{wrapfig}

\usepackage{arydshln}
\makeatletter
\def\adl@drawiv#1#2#3{
        \hskip.5\tabcolsep
        \xleaders#3{#2.5\@tempdimb #1{1}#2.5\@tempdimb}%
                #2\z@ plus1fil minus1fil\relax
        \hskip.5\tabcolsep}
\newcommand{\cdashlinelr}[1]{%
  \noalign{\vskip\aboverulesep
          \global\let\@dashdrawstore\adl@draw
          \global\let\ adl@draw\adl@drawiv}
  \cdashline{#1}
  \noalign{\global\let\adl@draw\@dashdrawstore
          \vskip\belowrulesep}}
\makeatother

\usepackage{tikz}
\usetikzlibrary{automata, arrows, bayesnet, bending}

\usepackage{tikzscale}

\usepackage{svg}

\copyrightyear{2026}
\acmYear{2026}
\setcopyright{rightsretained}
\acmConference[UMAP '26]{Proceedings of the 34th ACM Conference on User Modeling, Adaptation and Personalization}{June 8--11, 2026}{Gothenburg, Sweden}
\acmBooktitle{Proceedings of the 34th ACM Conference on User Modeling, Adaptation and Personalization (UMAP '26), June 8--11, 2026, Gothenburg, Sweden}

\begin{document}
\title{Sustained Impact of Agentic Personalisation in Marketing}
\subtitle{A Longitudinal Case Study}

\author{Olivier Jeunen}
\affiliation{
  \institution{aampe}
  \city{Antwerp}
  \country{Belgium}
}

\author{Eleanor Hanna}
\affiliation{
  \institution{aampe}
  \city{Raleigh}
  \state{NC}
  \country{USA}
}

\author{Schaun Wheeler}
\affiliation{
  \institution{aampe}
  \city{Cary}
  \state{NC}
  \country{USA}
}

\begin{abstract}
In consumer applications, Customer Relationship Management (CRM) has traditionally relied on the manual optimisation of static, rule-based messaging strategies.
While adaptive and autonomous learning systems offer the promise of scalable personalisation, it remains unclear to what extent ``human-in-the-loop'' oversight is required to sustain performance uplift over time.
This paper presents a longitudinal case study analysing a real-world consumer application that leverages agentic infrastructure to personalise marketing messaging for a large-scale user base over an 11-month period.

We compare two distinct periods: an \emph{active} phase where marketers directly curated content, audiences, and strategies---followed immediately by a \emph{passive} phase where agents operated autonomously from a fixed library of components.
Our results demonstrate that whilst active human management generates the highest relative lift in engagement metrics, the autonomous agents successfully sustained a positive lift during the passive period.
These findings suggest a symbiotic model where human intervention drives strategic initialisation and discovery, yet autonomous agents can ensure the scalable retention and preservation of performance gains.
\end{abstract}

\maketitle

\begin{figure*}[!t]
    \centering
    \includegraphics[width=\linewidth]{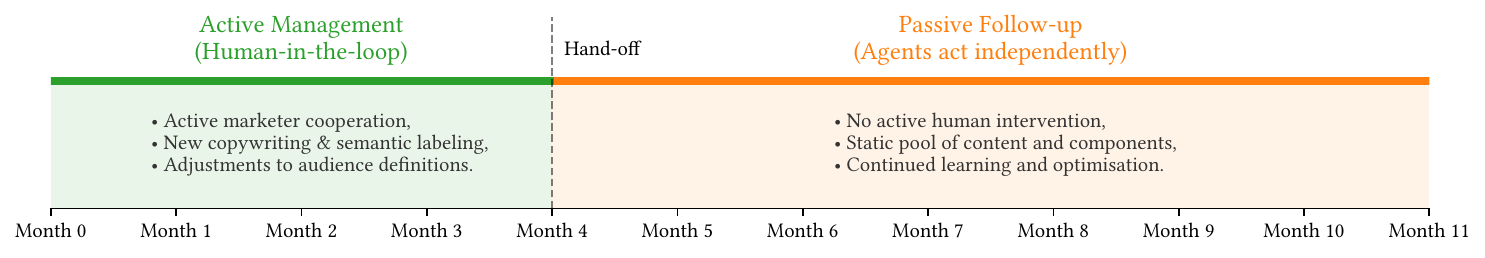}
    \caption{A timeline describing our longitudinal setup. We analyse 11 months of randomised controlled trial data, where the first 4 months represent an ``active management'' phase that benefits from dedicated marketing professionals that serve as humans-in-the-loop. The latter 7 months are a ``passive follow-up'' where agents act autonomously, whilst learning continuously.}
    \label{fig:timeline}
\end{figure*}

\section{Introduction \& Motivation}
Effective Customer Relationship Management (CRM) in consumer applications relies on the timely delivery of relevant content to re-engage users~\cite{Kumar2018}.
Traditionally, this orchestration has largely been a manual and iterative endeavour.
Marketing teams define static, rule-based heuristics---segmenting users by demographics or past behaviour and scheduling campaigns based on intuition or limited A/B-testing~\cite{Tynan1987}.
However, as user bases grow to the millions and content catalogues expand, this manual approach faces a scalability bottleneck.
The combinatorial complexity of matching the right user to the right message at the optimal time exceeds human processing capacity~\cite{GomezUribe2016}, leading to generic ``broadcast'' strategies that suffer from diminishing engagement returns whilst failing to deliver truly personalised user experiences.

To address this scaling challenge, the industry has increasingly turned to machine learning and autonomous agents to operationalise personalisation in marketing messaging.
The efficacy of this shift and widespread adoption among leading platforms has been well-documented: 
Duolingo~\cite{Yancey2020}, Kuaishou~\cite{Bie2025}, LinkedIn~\cite{Gupta2016, Gao2018, Prabhakar2022, Yuan2022}, Meta~\cite{Kroer2023},  Netflix~\cite{Huang2020},  Pinterest~\cite{Zhao2018},  Tencent~\cite{Zhang2023} and Twitter/X~\cite{Obrien2022} have all pursued this as an active research direction. % in the last decade.

Similarly, recent work by \citet{Abboud2025} highlights the potential of an \emph{agentic} framing, moving beyond the narrow view of personalisation as a \emph{prediction} problem, favouring a \emph{decision}-theoretic and autonomous lens~\cite{Joachims2021, CONSEQUENCES2022}.
These systems promise a transition from rigid, marketer-defined schedules to adaptive, fluid engagement strategies that learn continuously from user feedback.

Despite these technological advancements, a critical operational question remains: what is the long-term role of the human marketer in an agentic loop?
While the \emph{theoretical} capabilities of reinforcement learning and bandit algorithms suggest fully autonomous operation, industry practitioners often hesitate to remove human oversight entirely due to fears of model drift, feedback loops, or misalignment with brand strategy.
It remains unclear whether ``human-in-the-loop'' management is a constant necessity to prevent performance degradation, or if a more episodic intervention model is sufficient.
Furthermore, much of the existing literature focuses on short-term A/B-tests, leaving a gap in our understanding of how these systems behave longitudinally under real-world constraints.

Our work attempts to bridge this gap by presenting a longitudinal case study of a large-scale deployment of agentic CRM infrastructure.
We focus on a re-activation use-case where messaging targets users that have not engaged for a given amount of time---an audience comprising 8.8 million users.
We analyse an 11 month long randomised controlled trial, comparing a distinct phase of active marketer curation against a subsequent phase of autonomous agent operation---as visualised in Figure~\ref{fig:timeline}.
Our contributions are two-fold:
\begin{enumerate}
    \item We provide empirical evidence that whilst active human management drives the highest uplift in performance, autonomous agents are capable of effectively sustaining these gains over long periods without continuous intervention.
    \item We propose a symbiotic operational model for industry practitioners, defining clear roles for human creativity in strategic initialisation and algorithmic agents in scalable execution.
\end{enumerate}
\section{System Description \& Background}
\label{sec:system_background}
\begin{figure}[!t]
    \centering
    \includesvg[width=\linewidth]{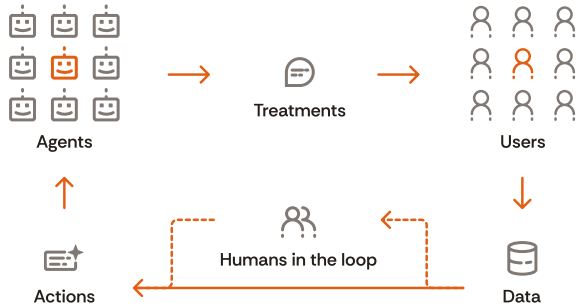}
    \caption{Visualising the feedback loop comprising end users, marketers, and agents in our setup.
    From ~\citet{Abboud2025}.}
    \label{fig:loop}
\end{figure}

The experimental platform utilised in this case study is based on the sequential decision-making framework detailed by \citet{Abboud2025}.
We refer the interested reader to their work for a full technical derivation, and briefly summarise the core components relevant to this longitudinal analysis.
Figure~\ref{fig:loop} summarises the feedback loop in which autonomous agents and marketing professionals operate.

\subsection{Sequential Decision-Making Formulation}
The system frames marketing orchestration as a sequential decision-making problem under uncertainty.
For a given user $u$ at time step $t$, the agent observes a state vector $s_{u,t}$ (encapsulating user demographics, past interactions, and app-usage history) and selects an action $a_t$ from a composite action space $\mathcal{A}$. This action is not a static template, but a dynamically assembled tuple comprising:

\begin{itemize}
    \item \textbf{Content Assembly:} the agent selects atomic content components (e.g. specific value propositions, greetings, visual assets) to construct a coherent message.
    \item \textbf{Timing Strategy:} the determination of the optimal send-time (e.g. specific hour of day, day of week, and frequency) relative to the user's habitual engagement patterns.
    \item \textbf{Channel Selection:} the optimal delivery medium (e.g. push notification, e-mail, SMS) to maximise reach and incremental engagement whilst minimising intrusion.
\end{itemize}

The agent's policy $\pi(a_t|s_{u,t})$ learns to maximise a cumulative reward signal $r_t$ (e.g. app opens, intent and conversion events)~\cite{Vasile2020}.
Specifically, the reward signal is a learnt scalarisation of various configurable events, augmented with a difference-in-differences design to measure incrementality~\cite{Athey2006}.
To balance the exploration of new strategies against the exploitation of known high-performers, the system leverages Thompson Sampling~\cite{Thompson1933}, maintaining a probabilistic distribution over the expected utility of potential actions.

\subsection{Human-Agent Interaction}
Crucially, the system is designed to scale through a modular content pipeline.
Rather than writing thousands of individual message variants, ``human-in-the-loop'' marketers provide the system with atomic components---such as distinct tones of voice, offer types, greetings, et cetera.
The agentic infrastructure then autonomously permutes these atoms to generate and serve personalised variants.
This approach has seen various empirical successes~\cite{Jeunen_ECIR2026,Abboud2025}.

In the context of this study, we treat the underlying algorithm as a fixed operational substrate.
Our analysis focuses specifically on the longitudinal efficacy of this system when the human supply and refinement of these atomic components is actively maintained, versus when it is paused.
\section{Experimental Setup}

\paragraph{Context \& Dataset}
The study was conducted in partnership with a leading multi-vertical consumer delivery application (anonymised for commercial confidentiality) serving a user base in the millions.
The platform offers a diverse catalogue ranging from restaurant delivery to grocery and retail goods.
The experiment utilised an industry standard randomized controlled trial design~\cite{kohavi2020trustworthy}.
Users were randomly assigned to either a treatment group, managed by the agentic infrastructure, or a holdout control group receiving the platform's standard business-as-usual (BAU) messaging logic.

This trial focused on a \emph{reactivation} use-case, where users were considered messageable if they had been inactive for a given amount of time.
This yielded a total audience of roughly 8.8 million users with a 90-10\% split across Agentic and BAU respectively.

\paragraph{Timeline \& Metrics}
The longitudinal analysis spans an 11-month period, divided into two distinct operational phases to evaluate the interplay between human strategy and autonomous execution.
Figure~\ref{fig:timeline} visualises this dichotomy between an \emph{active} and \emph{passive} phase.
We evaluate performance across the full conversion funnel, focusing on user retention.
Our primary metrics measure the number of days a user performs a given activity on the application---higher numbers not only reflect more activity, but also retention of returning users.
We include \textbf{Engagement} (any positive logged interaction), \textbf{Direct Opens} (attributed directly from clicks on push notifications), \textbf{Intent} (upper funnel events), and \textbf{Conversions} (lower funnel events).
We also monitor \textbf{Gross Merchandise Value} (GMV) to ensure that engagement and conversion lifts translate into economic impact.

\paragraph{Statistical Methodology}
To quantify the impact of the agentic system, we calculate the relative lift ($\Delta$) between the treatment and holdout groups.
Our longitudinal design, incorporating a persistent holdout group, effectively controls for temporal dynamics.
By measuring the $\Delta$ between treatment and control at each time step, we ensure that external factors---such as weekly periodicity and seasonal demand fluctuations---affect both groups equally and thus do not confound the performance attribution.
Given the high variance inherent in consumer behaviour data (attributable to sparse conversion events), we employ ML-RATE to reduce estimation variance whilst retaining unbiasedness in our average treatment effect estimates~\cite{Guo2021}.
The Delta method provides asymptotic variance estimates for the relative uplift estimand~\cite[Ch. 2]{Owen2013}.

\section{Results \& Discussion}

\begin{figure}[t]
    \centering
    \includegraphics[width=\linewidth]{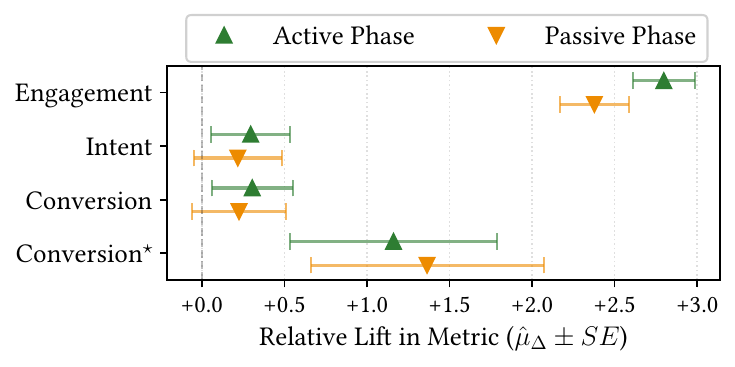}
    \caption{Relative lift across metrics and phases.}
    \label{fig:placeholder2}
\end{figure}

\paragraph{The Value of the Human-in-the-Loop}
\label{sec:active_results}
During the \emph{active} management phase, the influence from human marketers to guide autonomous agents yielded the largest incremental lift in performance metrics.
As shown in Figure~\ref{fig:placeholder2}, results are unilaterally positive across all funnel stages relative to control.

Not visualised but notably---we observed a +65.3\% ($\pm 1.31$) increase in days with direct app opens attributable to push notifications, validating the agents' ability to optimise composite messaging strategies effectively.
Overall days with positive engagement were up by +2.8\% ($\pm 0.19$), days with intent by +0.3\%  ($\pm 0.23$), and days with conversions by +0.3\%  ($\pm 0.24$).
Agentic messaging especially favoured conversions within a given vertical that is important to the business (\textbf{Conversion$^{\star}$}), up by $+1.2\% (\pm 0.62)$.

We have validated that these findings translate to proportional updates in GMV (i.e. no adverse effects on conversion magnitude).

\paragraph{Sustained Impact \& Robust Autonomy}
A key finding of this study is the system's resilience during the \emph{passive} phase.
Despite the cessation of manual content injection and optimisation for 7 months, performance across funnel metrics did not revert to the baseline expectation of the BAU experience.
Autonomous agents successfully sustained a +57\% $(\pm 0.11)$ lift in notification clicks, +2.4\% ($\pm 0.21$) days with engagement per user, and a +0.2\% ($\pm 0.27$) lift in intent and conversion metrics.
This validates that the infrastructure is not merely a delivery layer for human-managed message orchestration, but possesses an intrinsic ability to learn and adapt to user preferences over time, preventing the ``cliff-edge'' decay often seen in static rule-based campaigns.

\paragraph{Implications: A Symbiotic Model}
Whilst the passive phase confirms robustness, the delta between phases reveals the tangible value of human oversight.
The transition to fully autonomous operation resulted in a relative dampening of lift: roughly $-15$\% for overall engagement, $-12\%$ for notification opens,  and $-26\%$ for days with intent and days with conversion events per user---compared to the uplift observed during the active phase.
These results suggest a symbiotic operating model for industry practitioners:
%\begin{enumerate}
    %\item 
    
    \textbf{Strategic Initialisation}. Active human-in-the-loop management acts as a performance multiplier, necessary for driving initial lift, strategic discovery, and setting up a component library.
    
    %\item 
    \textbf{Autonomous Maintenance}. Agents serve as a high-fidelity retention layer, sustaining gains and ensuring that performance remains significantly above baseline even when human attention is diverted elsewhere.
    
%\end{enumerate}
We additionally studied the trend in relative uplift over time, observing a clear pattern across all metrics: a rising trend during the active management period, a plateauing trend during the first 4 months of the passive period, and a declining trend towards the end of the trial.
This reinforces our view that agents independently sustain performance improvements brought by an initial, human-led, strategic effort to set up a component library—for a period spanning several months. 
Nevertheless, active human-agent management and cooperation maximises potential and sustained impact.

\section{Conclusions \& Outlook}
Customer Relationship Management and marketing messaging are moving from rule-driven strategies to autonomous, adaptive, learned personalisation systems.
This shift brings along questions regarding the long-term feasibility and resilience of such systems under varying human-in-the-loop dynamics.

In this paper, we presented a longitudinal analysis of an agentic personalisation system deployed in a large-scale consumer application over an 11-month period.
By contrasting a phase of active human management against a subsequent phase of autonomous operation, we quantified both the operational resilience of agentic systems and the specific value-add of marketing professionals.

Our results challenge the dichotomy that CRM must be either fully manual or autonomous.
We propose a symbiotic model:

\textbf{Agents as the Baseline:} The agentic infrastructure successfully maintained positive lift (i.a. +57\% in direct opens) even after 7 months of independent operation, proving that autonomous systems can prevent a steep performance decay that is typically associated with static rule-based campaigns.

\textbf{Humans as the Multiplier:} Active human management provided a critical ``performance premium'' (an additional $\sim$12--26\% lift), suggesting that human intervention remains essential for strategic novelty and creative refreshment.

\paragraph{Future Outlook: Generative Augmentation}
Whilst this study leverages human-authored ``atomic content'' components, the natural evolution of this framework lies with generative models.
The current bottleneck---manual production of diverse message variants for the agents to experiment with---is a prime candidate for automation via Large Language Models (LLMs).

Nevertheless, these models come with their own sets of challenges regarding steerability, hallucinations, and consistency~\cite{Wheeler2025}.
Recent work has started to look into these applications areas, reporting operational successes~\cite{Bie2025}.
These systems could potentially help close the gap between the ``active'' and ``passive'' phases observed in this study, enabling the agent not just to \emph{select} the best available content components that are available, but to dynamically \emph{create} the optimal message for every user context.

\balance
%% References
\bibliographystyle{ACM-Reference-Format}
\bibliography{bibliography}

%%% -*-BibTeX-*-
%%% Do NOT edit. File created by BibTeX with style
%%% ACM-Reference-Format-Journals [18-Jan-2012].

\begin{thebibliography}{25}

%%% ====================================================================
%%% NOTE TO THE USER: you can override these defaults by providing
%%% customized versions of any of these macros before the \bibliography
%%% command.  Each of them MUST provide its own final punctuation,
%%% except for \shownote{} and \showURL{}.  The latter two
%%% do not use final punctuation, in order to avoid confusing it with
%%% the Web address.
%%%
%%% To suppress output of a particular field, define its macro to expand
%%% to an empty string, or better, \unskip, like this:
%%%
%%% \newcommand{\showURL}[1]{\unskip}   % LaTeX syntax
%%%
%%% \def \showURL #1{\unskip}           % plain TeX syntax
%%%
%%% ====================================================================

\ifx \showCODEN    \undefined \def \showCODEN     #1{\unskip}     \fi
\ifx \showISBNx    \undefined \def \showISBNx     #1{\unskip}     \fi
\ifx \showISBNxiii \undefined \def \showISBNxiii  #1{\unskip}     \fi
\ifx \showISSN     \undefined \def \showISSN      #1{\unskip}     \fi
\ifx \showLCCN     \undefined \def \showLCCN      #1{\unskip}     \fi
\ifx \shownote     \undefined \def \shownote      #1{#1}          \fi
\ifx \showarticletitle \undefined \def \showarticletitle #1{#1}   \fi
\ifx \showURL      \undefined \def \showURL       {\relax}        \fi
% The following commands are used for tagged output and should be
% invisible to TeX
\providecommand\bibfield[2]{#2}
\providecommand\bibinfo[2]{#2}
\providecommand\natexlab[1]{#1}
\providecommand\showeprint[2][]{arXiv:#2}

\bibitem[Abboud et~al\mbox{.}(2025)]%
        {Abboud2025}
\bibfield{author}{\bibinfo{person}{Sami Abboud}, \bibinfo{person}{Eleanor Hanna}, \bibinfo{person}{Olivier Jeunen}, \bibinfo{person}{Vineesha Raheja}, {and} \bibinfo{person}{Schaun Wheeler}.} \bibinfo{year}{2025}\natexlab{}.
\newblock \showarticletitle{Agentic Personalisation of Cross-Channel Marketing Experiences}. In \bibinfo{booktitle}{\emph{Proceedings of the Nineteenth ACM Conference on Recommender Systems}} \emph{(\bibinfo{series}{RecSys '25})}. \bibinfo{publisher}{ACM}, \bibinfo{pages}{907–910}.
\newblock
\href{https://doi.org/10.1145/3705328.3748125}{doi:\nolinkurl{10.1145/3705328.3748125}}


\bibitem[Athey and Imbens(2006)]%
        {Athey2006}
\bibfield{author}{\bibinfo{person}{Susan Athey} {and} \bibinfo{person}{Guido~W. Imbens}.} \bibinfo{year}{2006}\natexlab{}.
\newblock \showarticletitle{Identification and Inference in Nonlinear Difference-in-Differences Models}.
\newblock \bibinfo{journal}{\emph{Econometrica}} \bibinfo{volume}{74}, \bibinfo{number}{2} (\bibinfo{year}{2006}), \bibinfo{pages}{431--497}.
\newblock
\href{https://doi.org/10.1111/j.1468-0262.2006.00668.x}{doi:\nolinkurl{10.1111/j.1468-0262.2006.00668.x}}


\bibitem[Bie et~al\mbox{.}(2025)]%
        {Bie2025}
\bibfield{author}{\bibinfo{person}{Shifu Bie}, \bibinfo{person}{Jiangxia Cao}, \bibinfo{person}{Zixiao Luo}, \bibinfo{person}{Yichuan Zou}, \bibinfo{person}{Lei Liang}, \bibinfo{person}{Lu Zhang}, \bibinfo{person}{Linxun Chen}, \bibinfo{person}{Zhaojie Liu}, \bibinfo{person}{Xuanping Li}, \bibinfo{person}{Guorui Zhou}, \bibinfo{person}{Kaiqiao Zhan}, {and} \bibinfo{person}{Kun Gai}.} \bibinfo{year}{2025}\natexlab{}.
\newblock \bibinfo{title}{PushGen: Push Notifications Generation with LLM}.
\newblock
\showeprint[arxiv]{2512.14490}~[cs.IR]
\urldef\tempurl%
\url{https://arxiv.org/abs/2512.14490}
\showURL{%
\tempurl}
\newblock
\shownote{To appear in WSDM '26}.


\bibitem[Gao et~al\mbox{.}(2018)]%
        {Gao2018}
\bibfield{author}{\bibinfo{person}{Yan Gao}, \bibinfo{person}{Viral Gupta}, \bibinfo{person}{Jinyun Yan}, \bibinfo{person}{Changji Shi}, \bibinfo{person}{Zhongen Tao}, \bibinfo{person}{PJ Xiao}, \bibinfo{person}{Curtis Wang}, \bibinfo{person}{Shipeng Yu}, \bibinfo{person}{Romer Rosales}, \bibinfo{person}{Ajith Muralidharan}, {and} \bibinfo{person}{Shaunak Chatterjee}.} \bibinfo{year}{2018}\natexlab{}.
\newblock \showarticletitle{Near Real-time Optimization of Activity-based Notifications}. In \bibinfo{booktitle}{\emph{Proceedings of the 24th ACM SIGKDD International Conference on Knowledge Discovery \& Data Mining}} (London, United Kingdom) \emph{(\bibinfo{series}{KDD '18})}. \bibinfo{publisher}{Association for Computing Machinery}, \bibinfo{address}{New York, NY, USA}, \bibinfo{pages}{283–292}.
\newblock
\showISBNx{9781450355520}
\href{https://doi.org/10.1145/3219819.3219880}{doi:\nolinkurl{10.1145/3219819.3219880}}


\bibitem[Gomez-Uribe and Hunt(2016)]%
        {GomezUribe2016}
\bibfield{author}{\bibinfo{person}{Carlos~A. Gomez-Uribe} {and} \bibinfo{person}{Neil Hunt}.} \bibinfo{year}{2016}\natexlab{}.
\newblock \showarticletitle{The Netflix Recommender System: Algorithms, Business Value, and Innovation}.
\newblock \bibinfo{journal}{\emph{ACM Trans. Manage. Inf. Syst.}} \bibinfo{volume}{6}, \bibinfo{number}{4}, Article \bibinfo{articleno}{13} (\bibinfo{date}{Dec.} \bibinfo{year}{2016}), \bibinfo{numpages}{19}~pages.
\newblock
\showISSN{2158-656X}
\href{https://doi.org/10.1145/2843948}{doi:\nolinkurl{10.1145/2843948}}


\bibitem[Guo et~al\mbox{.}(2021)]%
        {Guo2021}
\bibfield{author}{\bibinfo{person}{Yongyi Guo}, \bibinfo{person}{Dominic Coey}, \bibinfo{person}{Mikael Konutgan}, \bibinfo{person}{Wenting Li}, \bibinfo{person}{Chris Schoener}, {and} \bibinfo{person}{Matt Goldman}.} \bibinfo{year}{2021}\natexlab{}.
\newblock \showarticletitle{Machine Learning for Variance Reduction in Online Experiments}. In \bibinfo{booktitle}{\emph{Advances in Neural Information Processing Systems}}, Vol.~\bibinfo{volume}{34}. \bibinfo{publisher}{Curran Associates, Inc.}, \bibinfo{pages}{8637--8648}.
\newblock


\bibitem[Gupta et~al\mbox{.}(2016)]%
        {Gupta2016}
\bibfield{author}{\bibinfo{person}{Rupesh Gupta}, \bibinfo{person}{Guanfeng Liang}, \bibinfo{person}{Hsiao-Ping Tseng}, \bibinfo{person}{Ravi~Kiran Holur~Vijay}, \bibinfo{person}{Xiaoyu Chen}, {and} \bibinfo{person}{Romer Rosales}.} \bibinfo{year}{2016}\natexlab{}.
\newblock \showarticletitle{Email Volume Optimization at LinkedIn}. In \bibinfo{booktitle}{\emph{Proceedings of the 22nd ACM SIGKDD International Conference on Knowledge Discovery and Data Mining}} (San Francisco, California, USA) \emph{(\bibinfo{series}{KDD '16})}. \bibinfo{publisher}{Association for Computing Machinery}, \bibinfo{address}{New York, NY, USA}, \bibinfo{pages}{97–106}.
\newblock
\showISBNx{9781450342322}
\href{https://doi.org/10.1145/2939672.2939692}{doi:\nolinkurl{10.1145/2939672.2939692}}


\bibitem[Huang(2020)]%
        {Huang2020}
\bibfield{author}{\bibinfo{person}{Grace Huang}.} \bibinfo{year}{2020}\natexlab{}.
\newblock \bibinfo{title}{Mesa: Building a Personalized Messaging System at Netflix}.
\newblock \bibinfo{howpublished}{Data Council}.
\newblock
\urldef\tempurl%
\url{https://aicouncil.com/talks/mesa-building-a-personalized-messaging-system-at-netflix}
\showURL{%
\tempurl}
\newblock
\shownote{Accessed: 2026-02}.


\bibitem[Jeunen et~al\mbox{.}(2022)]%
        {CONSEQUENCES2022}
\bibfield{author}{\bibinfo{person}{Olivier Jeunen}, \bibinfo{person}{Thorsten Joachims}, \bibinfo{person}{Harrie Oosterhuis}, \bibinfo{person}{Yuta Saito}, {and} \bibinfo{person}{Flavian Vasile}.} \bibinfo{year}{2022}\natexlab{}.
\newblock \showarticletitle{CONSEQUENCES — Causality, Counterfactuals and Sequential Decision-Making for Recommender Systems}. In \bibinfo{booktitle}{\emph{Proc. of the 16th ACM Conference on Recommender Systems}} \emph{(\bibinfo{series}{RecSys '22})}. \bibinfo{publisher}{ACM}, \bibinfo{pages}{654–657}.
\newblock
\showISBNx{9781450392785}
\href{https://doi.org/10.1145/3523227.3547409}{doi:\nolinkurl{10.1145/3523227.3547409}}


\bibitem[Jeunen and Wheeler(2026)]%
        {Jeunen_ECIR2026}
\bibfield{author}{\bibinfo{person}{Olivier Jeunen} {and} \bibinfo{person}{Schaun Wheeler}.} \bibinfo{year}{2026}\natexlab{}.
\newblock \showarticletitle{Behavioural Effects of Agentic Messaging}. In \bibinfo{booktitle}{\emph{Advances in Information Retrieval}}. \bibinfo{publisher}{Springer Nature Switzerland}, \bibinfo{pages}{105--110}.
\newblock
\showISBNx{978-3-032-21321-1}


\bibitem[Joachims et~al\mbox{.}(2021)]%
        {Joachims2021}
\bibfield{author}{\bibinfo{person}{Thorsten Joachims}, \bibinfo{person}{Ben London}, \bibinfo{person}{Yi Su}, \bibinfo{person}{Adith Swaminathan}, {and} \bibinfo{person}{Lequn Wang}.} \bibinfo{year}{2021}\natexlab{}.
\newblock \showarticletitle{Recommendations as Treatments}.
\newblock \bibinfo{journal}{\emph{AI Magazine}} \bibinfo{volume}{42}, \bibinfo{number}{3} (\bibinfo{date}{Nov.} \bibinfo{year}{2021}), \bibinfo{pages}{19--30}.
\newblock


\bibitem[Kohavi et~al\mbox{.}(2020)]%
        {kohavi2020trustworthy}
\bibfield{author}{\bibinfo{person}{Ron Kohavi}, \bibinfo{person}{Diane Tang}, {and} \bibinfo{person}{Ya Xu}.} \bibinfo{year}{2020}\natexlab{}.
\newblock \bibinfo{booktitle}{\emph{Trustworthy online controlled experiments: A practical guide to A/B testing}}.
\newblock \bibinfo{publisher}{Cambridge University Press}.
\newblock


\bibitem[Kroer et~al\mbox{.}(2023)]%
        {Kroer2023}
\bibfield{author}{\bibinfo{person}{Christian Kroer}, \bibinfo{person}{Deeksha Sinha}, \bibinfo{person}{Xuan Zhang}, \bibinfo{person}{Shiwen Cheng}, {and} \bibinfo{person}{Ziyu Zhou}.} \bibinfo{year}{2023}\natexlab{}.
\newblock \bibinfo{title}{Fair Notification Optimization: An Auction Approach}.
\newblock
\showeprint[arxiv]{2302.04835}~[cs.GT]
\urldef\tempurl%
\url{https://arxiv.org/abs/2302.04835}
\showURL{%
\tempurl}


\bibitem[Kumar and Reinartz(2018)]%
        {Kumar2018}
\bibfield{author}{\bibinfo{person}{Vineet Kumar} {and} \bibinfo{person}{Werner Reinartz}.} \bibinfo{year}{2018}\natexlab{}.
\newblock \bibinfo{booktitle}{\emph{Customer relationship management}}.
\newblock \bibinfo{publisher}{Springer}.
\newblock


\bibitem[O'Brien et~al\mbox{.}(2022)]%
        {Obrien2022}
\bibfield{author}{\bibinfo{person}{Conor O'Brien}, \bibinfo{person}{Huasen Wu}, \bibinfo{person}{Shaodan Zhai}, \bibinfo{person}{Dalin Guo}, \bibinfo{person}{Wenzhe Shi}, {and} \bibinfo{person}{Jonathan~J Hunt}.} \bibinfo{year}{2022}\natexlab{}.
\newblock \bibinfo{title}{Should I send this notification? Optimizing push notifications decision making by modeling the future}.
\newblock
\showeprint[arxiv]{2202.08812}~[cs.IR]
\urldef\tempurl%
\url{https://arxiv.org/abs/2202.08812}
\showURL{%
\tempurl}


\bibitem[Owen(2013)]%
        {Owen2013}
\bibfield{author}{\bibinfo{person}{Art~B. Owen}.} \bibinfo{year}{2013}\natexlab{}.
\newblock \bibinfo{booktitle}{\emph{Monte Carlo theory, methods and examples}}.
\newblock


\bibitem[Prabhakar et~al\mbox{.}(2022)]%
        {Prabhakar2022}
\bibfield{author}{\bibinfo{person}{Prakruthi Prabhakar}, \bibinfo{person}{Yiping Yuan}, \bibinfo{person}{Guangyu Yang}, \bibinfo{person}{Wensheng Sun}, {and} \bibinfo{person}{Ajith Muralidharan}.} \bibinfo{year}{2022}\natexlab{}.
\newblock \showarticletitle{Multi-objective Optimization of Notifications Using Offline Reinforcement Learning}. In \bibinfo{booktitle}{\emph{Proceedings of the 28th ACM SIGKDD Conference on Knowledge Discovery and Data Mining}} (Washington DC, USA) \emph{(\bibinfo{series}{KDD '22})}. \bibinfo{publisher}{Association for Computing Machinery}, \bibinfo{address}{New York, NY, USA}, \bibinfo{pages}{3752–3760}.
\newblock
\showISBNx{9781450393850}
\href{https://doi.org/10.1145/3534678.3539193}{doi:\nolinkurl{10.1145/3534678.3539193}}


\bibitem[Thompson(1933)]%
        {Thompson1933}
\bibfield{author}{\bibinfo{person}{William~R. Thompson}.} \bibinfo{year}{1933}\natexlab{}.
\newblock \showarticletitle{On the Likelihood that One Unknown Probability Exceeds Another in View of the Evidence of Two Samples}.
\newblock \bibinfo{journal}{\emph{Biometrika}} \bibinfo{volume}{25}, \bibinfo{number}{3/4} (\bibinfo{year}{1933}), \bibinfo{pages}{285--294}.
\newblock
\showISSN{00063444}
\urldef\tempurl%
\url{http://www.jstor.org/stable/2332286}
\showURL{%
\tempurl}


\bibitem[Tynan and Drayton(1987)]%
        {Tynan1987}
\bibfield{author}{\bibinfo{person}{A.~Caroline Tynan} {and} \bibinfo{person}{Jennifer Drayton}.} \bibinfo{year}{1987}\natexlab{}.
\newblock \showarticletitle{Market segmentation}.
\newblock \bibinfo{journal}{\emph{Journal of Marketing Management}} \bibinfo{volume}{2}, \bibinfo{number}{3} (\bibinfo{year}{1987}), \bibinfo{pages}{301--335}.
\newblock
\href{https://doi.org/10.1080/0267257X.1987.9964020}{doi:\nolinkurl{10.1080/0267257X.1987.9964020}}


\bibitem[Vasile et~al\mbox{.}(2020)]%
        {Vasile2020}
\bibfield{author}{\bibinfo{person}{Flavian Vasile}, \bibinfo{person}{David Rohde}, \bibinfo{person}{Olivier Jeunen}, {and} \bibinfo{person}{Amine Benhalloum}.} \bibinfo{year}{2020}\natexlab{}.
\newblock \showarticletitle{A Gentle Introduction to Recommendation as Counterfactual Policy Learning}. In \bibinfo{booktitle}{\emph{Proc. of the 28th ACM Conference on User Modeling, Adaptation and Personalization}} \emph{(\bibinfo{series}{UMAP '20})}. \bibinfo{publisher}{ACM}, \bibinfo{pages}{392–393}.
\newblock
\showISBNx{9781450368612}
\href{https://doi.org/10.1145/3340631.3398666}{doi:\nolinkurl{10.1145/3340631.3398666}}


\bibitem[Wheeler and Jeunen(2025)]%
        {Wheeler2025}
\bibfield{author}{\bibinfo{person}{Schaun Wheeler} {and} \bibinfo{person}{Olivier Jeunen}.} \bibinfo{year}{2025}\natexlab{}.
\newblock \showarticletitle{Procedural Memory Is Not All You Need: Bridging Cognitive Gaps in LLM-Based Agents}. In \bibinfo{booktitle}{\emph{Adjunct Proceedings of the 33rd ACM Conference on User Modeling, Adaptation and Personalization}} \emph{(\bibinfo{series}{UMAP Adjunct '25})}. \bibinfo{publisher}{Association for Computing Machinery}, \bibinfo{address}{New York, NY, USA}, \bibinfo{pages}{360–364}.
\newblock
\showISBNx{9798400713996}
\href{https://doi.org/10.1145/3708319.3734172}{doi:\nolinkurl{10.1145/3708319.3734172}}


\bibitem[Yancey and Settles(2020)]%
        {Yancey2020}
\bibfield{author}{\bibinfo{person}{Kevin~P. Yancey} {and} \bibinfo{person}{Burr Settles}.} \bibinfo{year}{2020}\natexlab{}.
\newblock \showarticletitle{A Sleeping, Recovering Bandit Algorithm for Optimizing Recurring Notifications}. In \bibinfo{booktitle}{\emph{Proceedings of the 26th ACM SIGKDD International Conference on Knowledge Discovery \& Data Mining}} (Virtual Event, CA, USA) \emph{(\bibinfo{series}{KDD '20})}. \bibinfo{publisher}{Association for Computing Machinery}, \bibinfo{address}{New York, NY, USA}, \bibinfo{pages}{3008–3016}.
\newblock
\showISBNx{9781450379984}
\href{https://doi.org/10.1145/3394486.3403351}{doi:\nolinkurl{10.1145/3394486.3403351}}


\bibitem[Yuan et~al\mbox{.}(2022)]%
        {Yuan2022}
\bibfield{author}{\bibinfo{person}{Yiping Yuan}, \bibinfo{person}{Ajith Muralidharan}, \bibinfo{person}{Preetam Nandy}, \bibinfo{person}{Miao Cheng}, {and} \bibinfo{person}{Prakruthi Prabhakar}.} \bibinfo{year}{2022}\natexlab{}.
\newblock \showarticletitle{Offline Reinforcement Learning for Mobile Notifications}. In \bibinfo{booktitle}{\emph{Proceedings of the 31st ACM International Conference on Information \& Knowledge Management}} (Atlanta, GA, USA) \emph{(\bibinfo{series}{CIKM '22})}. \bibinfo{publisher}{Association for Computing Machinery}, \bibinfo{address}{New York, NY, USA}, \bibinfo{pages}{3614–3623}.
\newblock
\showISBNx{9781450392365}
\href{https://doi.org/10.1145/3511808.3557083}{doi:\nolinkurl{10.1145/3511808.3557083}}


\bibitem[Zhang et~al\mbox{.}(2023)]%
        {Zhang2023}
\bibfield{author}{\bibinfo{person}{Yuchen Zhang}, \bibinfo{person}{Mingjun Zhao}, \bibinfo{person}{Chenglin Li}, \bibinfo{person}{Weiyu Tou}, \bibinfo{person}{Haolan Chen}, \bibinfo{person}{Di Niu}, \bibinfo{person}{Cunxiang Yin}, \bibinfo{person}{Yancheng He}, {and} \bibinfo{person}{Fei Guo}.} \bibinfo{year}{2023}\natexlab{}.
\newblock \showarticletitle{Online Volume Optimization for Notifications via Long Short-Term Value Modeling}. In \bibinfo{booktitle}{\emph{Advances in Knowledge Discovery and Data Mining}}, \bibfield{editor}{\bibinfo{person}{Hisashi Kashima}, \bibinfo{person}{Tsuyoshi Ide}, {and} \bibinfo{person}{Wen-Chih Peng}} (Eds.). \bibinfo{publisher}{Springer Nature Switzerland}, \bibinfo{address}{Cham}, \bibinfo{pages}{16--28}.
\newblock
\showISBNx{978-3-031-33380-4}


\bibitem[Zhao et~al\mbox{.}(2018)]%
        {Zhao2018}
\bibfield{author}{\bibinfo{person}{Bo Zhao}, \bibinfo{person}{Koichiro Narita}, \bibinfo{person}{Burkay Orten}, {and} \bibinfo{person}{John Egan}.} \bibinfo{year}{2018}\natexlab{}.
\newblock \showarticletitle{Notification Volume Control and Optimization System at Pinterest}. In \bibinfo{booktitle}{\emph{Proceedings of the 24th ACM SIGKDD International Conference on Knowledge Discovery \& Data Mining}} (London, United Kingdom) \emph{(\bibinfo{series}{KDD '18})}. \bibinfo{publisher}{Association for Computing Machinery}, \bibinfo{address}{New York, NY, USA}, \bibinfo{pages}{1012–1020}.
\newblock
\showISBNx{9781450355520}
\href{https://doi.org/10.1145/3219819.3219906}{doi:\nolinkurl{10.1145/3219819.3219906}}


\end{thebibliography}

\end{document}